\documentclass[letterpaper, 10 pt, conference]{ieeeconf}  

\usepackage{graphicx}
\usepackage{color}
\usepackage{booktabs}
\usepackage{stfloats}
\usepackage{amssymb}

\usepackage{amsmath}
\usepackage{cite}
\usepackage{fancyhdr}

\IEEEoverridecommandlockouts                              

\title{\LARGE \bf
Abnormal Occupancy Grid Map Recognition using Attention Network
}
\author{Fuqin Deng$^{1,3,4\ast}$, Hua Feng$^{1\ast}$, Mingjian Liang$^{1}$, Qi Feng$^{1}$, Ningbo Yi$^{1}$,
\\Yong Yang$^{4}$, Yuan Gao$^{3}$, Junfeng Chen$^{3}$, and Tin Lun Lam$^{2,3,\dagger}$
\thanks{This work was supported in part by the National Key R$\&$D Program of China (2020YFB1313300), the funding (AC01202101025, AC01202101026) from the Shenzhen Institute of Artificial Intelligence and Robotics for Society, the special projects in key fields of Guangdong Provincial Department of Education (2019KZDZX1025), Guangdong Science and Technology Major Special Fund (No.2019-252), Innovative Program for Graduate Education (503170060259) from the Wuyi University, and Shenzhen Peacock Plan of Shenzhen Science and Technology Program (KQTD2016113010470345). We would also like to thank Dr. Li Nan Nan from Macau University of Science and Technology for his valuable discussion on this paper.}

\thanks{$^{\ast}$Authors contributed equally}
\thanks{$^{1}$School of Intelligent Manufacturing, the Wuyi University, Jiangmen, China.
        }%
\thanks{$^{2}$School of Science and Engineering, the Chinese University of Hong Kong, Shenzhen, China.
        }%
\thanks{$^{3}$The Shenzhen Institute of Artificial Intelligence and Robotics for Society, the Chinese University of Hong Kong, Shenzhen, China.
        }%
\thanks{$^{4}$The 3irobotix Co.,Ltd, Shenzhen, China.
        }%
\thanks{$^{\dagger}$Corresponding author is Tin Lun Lam
        {\tt\small tllam@cuhk.edu.cn}
        }%
}

\begin{document}

\maketitle
\thispagestyle{fancy}
\pagestyle{fancy}

\lhead{} 
\chead{} 
\rhead{} 
\lfoot{} 
\cfoot{} 
\rfoot{} 
\renewcommand{\headrulewidth}{0pt} 
\renewcommand{\footrulewidth}{0pt} 

\begin{abstract}

The occupancy grid map is a critical component of autonomous positioning and navigation in the mobile robotic system, as many other systems' performance depends heavily on it. To guarantee the quality of the occupancy grid maps, researchers previously had to perform tedious manual recognition for a long time. This work focuses on automatic abnormal occupancy grid map recognition using the residual neural networks and a novel attention mechanism module. We propose an effective channel and spatial Residual SE(csRSE) attention module, which contains a residual block for producing hierarchical features, followed by both channel SE (cSE) block and spatial SE (sSE) block for the sufficient information extraction along the channel and spatial pathways. To further summarize the occupancy grid map characteristics and experiment with our csRSE attention modules, we constructed a dataset called occupancy grid map dataset (OGMD) for our experiments. On this OGMD test dataset, we tested few variants of our proposed structure and compared them with other attention mechanisms. Our experimental results show that the proposed attention network can infer the abnormal map with state-of-the-art (SOTA) accuracy of 96.23\% for abnormal occupancy grid map recognition.

\end{abstract}

\section{INTRODUCTION}

The occupancy grid map was first introduced for surface point positions with two-dimensional (2D) planar grids \cite{elfes1989using}, which had gained great success fusing raw sensor data in one environment representation \cite{hachour2008path}. In the narrow indoor environments or spacious outdoor environments, occupancy grid map can be used for the autonomous positioning and navigation by collecting the position information of obstacles. In recent year, occupancy grid map has applications in obstacle avoidance \cite{borenstein1991vector}, multi-sensor data fusion \cite{dieterfox1998map}, object tracking \cite{wang2007simultaneous}, simultaneous localization and mapping (SLAM) \cite{hess2016real}, and multi-robot global localization \cite{guo2021semantic}.

\begin{figure}[t!]
    \centering
    \includegraphics [width=3.4in]{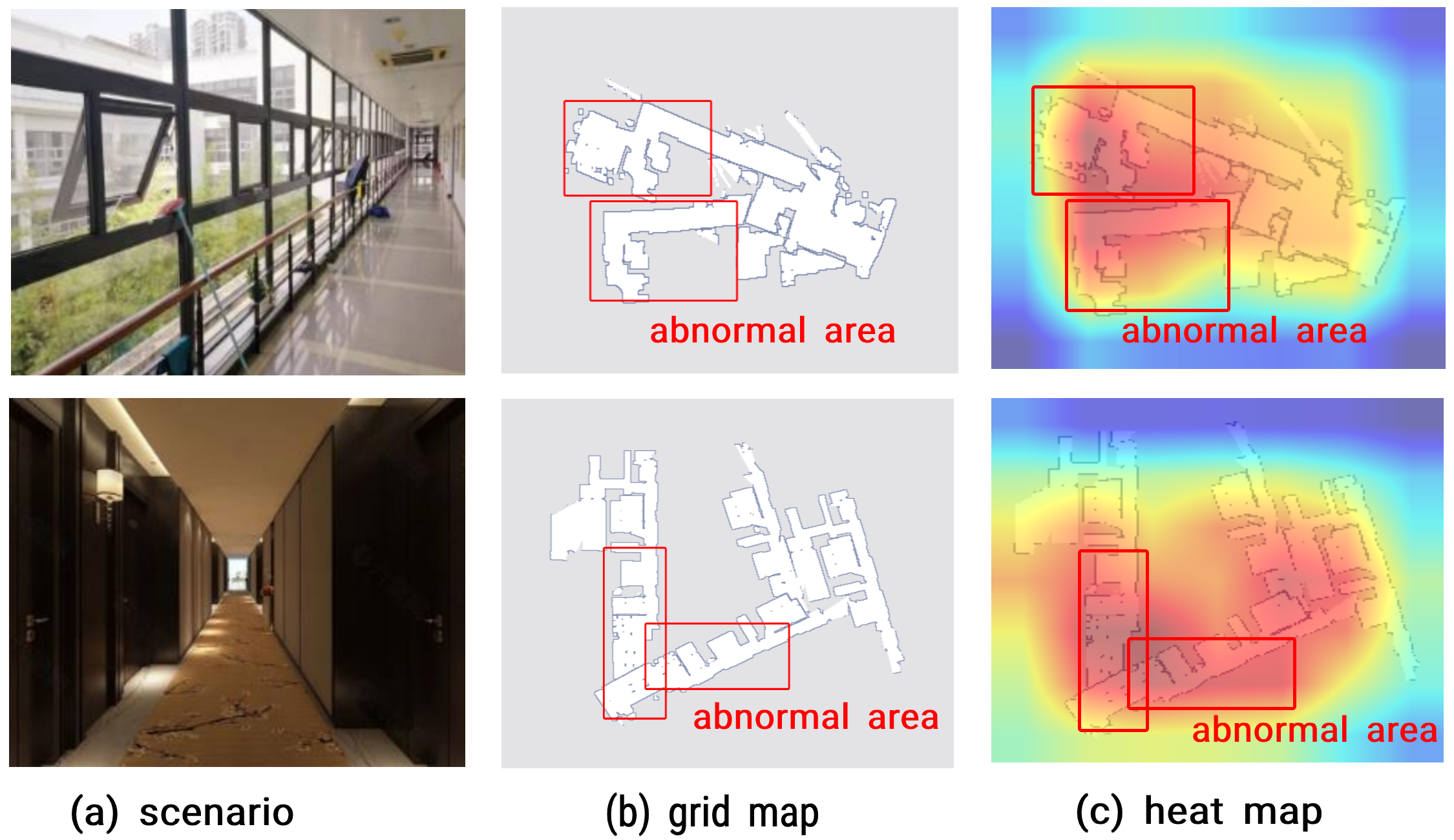}
    \vspace{-3mm}
    \caption{Problems with occupancy grid map in existing computer vision tasks. Column (a) represent domain-specific scenarios, (b) show the corresponding abnormal occupancy grid maps, and (c) represent attention heat map. In column (b), the regions marked in the red frame represent the abnormal areas. In column (c), our attention network method can detect these abnormal areas and then report these failure cases. More examples are in Fig.~\ref{visualization}}
    \label{examples}
    \vspace{-5mm}
\end{figure}

On the path to autonomous navigation, an occupancy grid map is essential for trajectory planning module \cite{ikeda2013indoor}. In the SLAM process, the occupancy grid maps aim at estimating the occupancy probability and recording discretized object's locations \cite{sodhi2019online}. In practice, many factors may affect the quality of the obtained occupancy grid maps. As shown in the $1st$ row of Fig.~\ref{examples} (a), when the mobile robot meets the transparent glass walls, the laser sensor cannot receive the laser light penetrating through the transparent glass, making it hard to determine the specific location of walls. Moreover, if the mobile robots without a loop detection module \cite{tzeng2006loop} builds an occupancy grid map on a long corridor in the $2nd$ row of Fig.~\ref{examples} (a), they may lose some critical scan measurement data. Due to these overlapped or incomplete occupancy grid maps (e. g. Fig.~\ref{examples} (b)), it would make mobile robots hard to plan the path and avoid obstacles during navigation.

For abnormal occupancy grid map recognition, there are mainly three types of methods, including shape estimation~\cite{tanzmeister2016evidential, roos2016reliable}, edge detection~\cite{scheel2018tracking} and data-driven methods~\cite{piewak2017fully}. The shape-based and edge-based methods are designed to associate measurements to object's feature, while some abnormal occupancy grid maps are falsely detected as positive. For data-driven method, especially with the convolutional neural networks (CNN) \cite{nielsen2015neural}, the abnormal occupancy grid map can be detected in terms of low-level features (e. g. blur, burr, and overlap). The recognition experiments with mainstream CNN-based method can get the approximately 91\% accuracy result (as shown in Tab.~\ref{accuracy}, e. g. ResNet32). Furthermore, based on attention mechanism ~\cite{olshausen1993neurobiological}, Squeeze-Excitation (SE) block ~\cite{hu2018squeeze} and convolutional block attention module (CBAM) \cite{woo2018cbam} have been embedded to the input feature map for adaptive feature refinement from channel-wise and spatial axes. Specially, these attention networks (with SE and CBAM) can achieve an accuracy result of approximately 94\% (as shown in Tab.~\ref{accuracy}, e. g. ResNet32 + SE). However, these methods are inefficient in terms of global feature and multi-scale feature extraction.

To improve representation ability of feature extraction, an effective network with the attention module is designed to aggregate specific features of occupancy grid map. Different from the SE and CBAM, we design a channel and spatial Residual SE(csRSE) attention module for the global context extraction on the different aggregation strategy. The proposed csRSE module is a global context modeling module which aggregates the features by using residual block and hybrid attention mechanisms. In csRSE module, the residual block can generate the hierarchical features and benefit gradient propagation. Besides, the cSE block and sSE block automatically contain sufficient spatial information of the intermediate layers and assign different attention weights to the abnormal regions. In our attention network, csRSE attention modules are embedded to generally focus on the multi-scale features and suppress unnecessary information. As in Fig.~\ref{examples} (c), the visualizing results of our csRSE module show the importance weights feature in occupancy grid maps. 

Main contributions of this work are listed as follows:
\begin{itemize}
\item We introduce a channel and spatial Residual SE (csRSE) attention module for global contextual extraction.
\item We contribute an occupancy grid map dataset (OGMD) for SLAM occupancy grid map recognition task.
\item We propose an attention network with multi csRSE modules for SOTA abnormal occupancy grid map recognition.
\end{itemize}


\section{RELATED WORK}

\subsection{Occupancy Grid Map}

The research on automatic map construction using robots has always been a key point for researchers \cite{matikainen2007classification, bertozzi2008obstacle}. Elfes et al. \cite{munz2009sensor} introduced a grid-based algorithm for 2D environment modeling. The continuous spaces of the environment are discretized by these evenly-spaced grids. According to the probabilistic formulation, each grid cell of constructed map represents the probability of the corresponding region that may be occupied, free or even not yet explored. With the help of occupancy grid map, the robot can identify the presence or absence of an obstacle in the space of the environment.

However, the occupancy grid map suffers from corrupted obstacle silhouettes, occlusions, and false distance estimates in static environments. To address these problems, a highly engineered method \cite{tanzmeister2016evidential} was proposed to extract information for the environment modeling. A data-driven method for map recognition is proposed in \cite{scheel2018tracking}, where the radar sensor measurement model was trained and this method outperforms the manually designed model. Piewak et al. \cite{piewak2017fully} trained a neural network to reduce false distance estimation in a dynamic environment. Their approach referred to a pixel-wise classification task to determine whether a cell in the actual environment is occupied or free.

\subsection{Deep Architectures and Attention Mechanisms}
In the last years, many attempts have been made to improve the original CNN architecture to achieve better accuracy. In image classification and object detection, recent approaches like AlexNet \cite{krizhevsky2012imagenet}, VGG-16 \cite{simonyan2014very}, InceptionNet \cite{szegedy2015going}, or MobileNet \cite{howard2017mobilenets} are based on the plain CNN architecture. However, CNN-based networks have problems in gradient propagation and convergence with the amount of data increases \cite{philipp2018gradients}. To address this issue, ResNet \cite{he2016deep} proposes a simple identity skip-connection to ease the optimization problem of deep networks. Our attention network is a backbone architecture based on ResNet, which can enhance the input features but also well benefit gradient propagation on occupancy grid map data training process.

Recently, various attention mechanisms had been proposed for image recognition tasks. Hu et al. \cite{hu2018squeeze} had proposed a compact SE block to extract the inter-channel information. Residual Attention Network (RAN) \cite{wang2017residual} modified ResNet by stacking identical soft attention modules which is beneficial to refine the feature maps. Spatial attention \cite{jaderberg2015spatial} had been made to recalibrate the channel dependency as an effective extraction module. Then, Woo et al. \cite{woo2018cbam} introduced a CBAM module that sequentially recalibrates channel and spatial attention to refine intermediate feature maps. However, these methods miss the global spatial information, which is an important factor for feature fusion and accurate attention map generation. Inspired by the global context block\cite{cao2019gcnet}, our csRSE attention network, which contains a residual block for producing hierarchical features, followed by both cSE block and sSE block for the sufficient spatial information extraction and the resulted hierarchical features enhancement.

\section{A NEW SLAM DATASET}
To facilitate the research of the occupancy grid map recognition problem, we contribute a new SLAM dataset containing 6916 occupancy grid maps through an indoor robot vacuum cleaner. To the best of our knowledge, OGMD is a benchmark specifically for the occupancy grid map recognition. We contribute the source code and our dataset public available\footnote{The available online website: https://github.com/ThomerShen/OGMD} to simulate future research.

\begin{figure}[ht]
    \centering
    \includegraphics [width=3.5in]{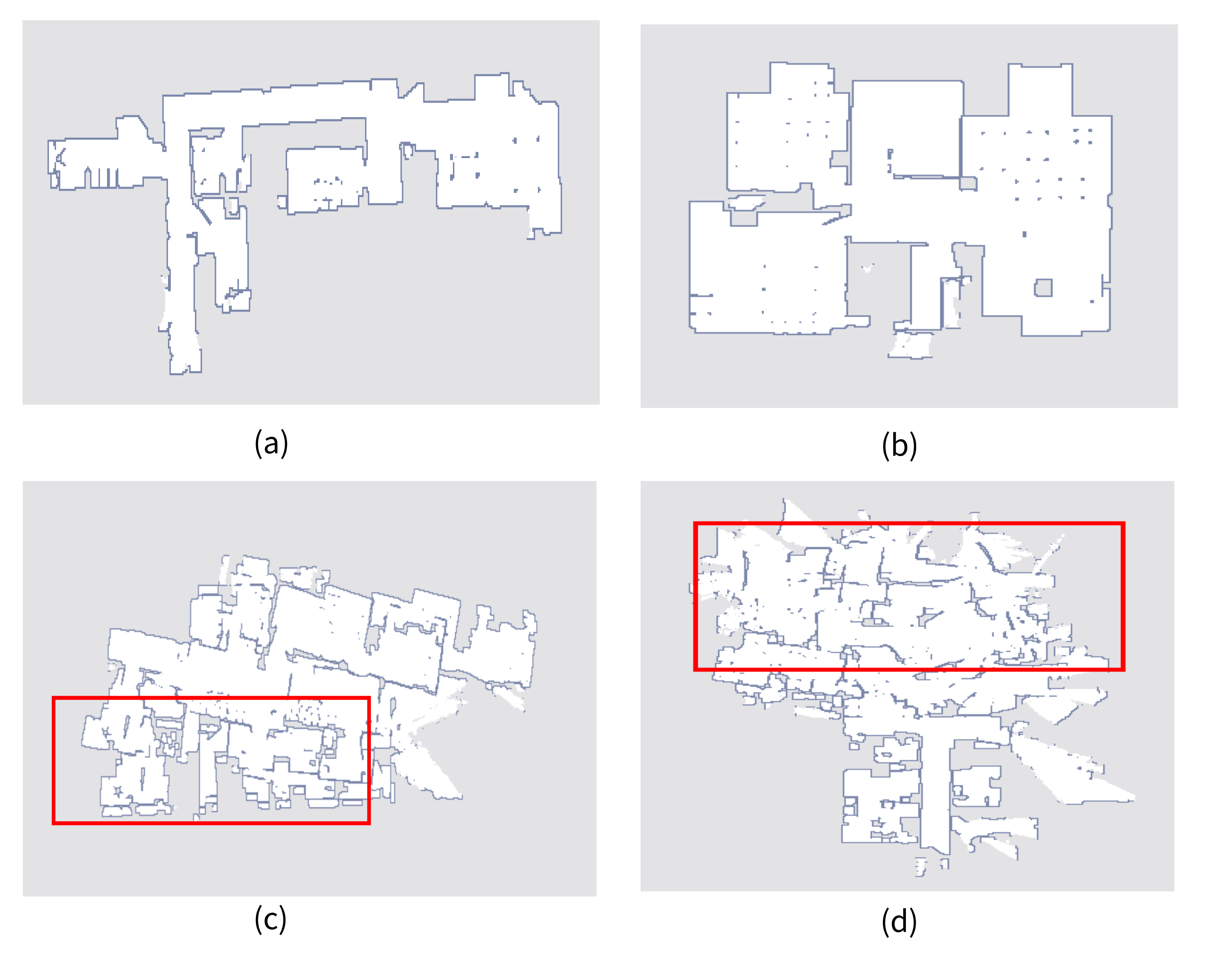}
    \vspace{-5mm}
    \caption{Example 2D occupancy grid maps in our OGMD. (a) and (b) represent normal occupancy grid maps, (c) and (d) represent abnormal occupancy grid maps. The regions marked in the red frame represent the abnormalities.}
    \label{dataset}
    \vspace{-3mm}
\end{figure}
\subsection{Dataset Construction}
The constructed OGMD covers diverse daily-life indoor environment scenarios, such as residential houses, super-markets, office buildings, hotels, and schools. These occupancy grid maps are created with an initial size of 50m×50m. To further increase the number of training examples, we applied random rotation and offset to cropped areas of 34m×34m used as training examples. These occupancy grid maps are labeled as two categories, including 3210 normal occupancy grid maps and 3706 abnormal occupancy grid maps respectively. In the following experiments, 6916 occupancy grid maps are randomly divided into 4150 training examples, 2079 validation examples, and 690 test examples, and the ratio is roughly 6:3:1.

\subsection{Dataset Analysis}
In OGMD, the occupancy grid maps are generated by the scan data of the robot laser sensor. For detail, each cell of occupancy grid map is obtained by the scan measurement data. In real collection scene, the occupancy grid maps are created by using either one scan or an accumulation of multiple sensor scans. The occupancy grid maps created by mobile robot are depicted in Fig.~\ref{dataset}.

Through the scans of the laser sensor, all obstacles are displayed as a line segment in the occupancy grid map. The aggregation of the scans represents the feature parameters of obstacles in the indoor environment. As shown in Fig.~\ref{dataset} (a) and (b), the bright white region of the occupancy grid map indicates a flat and open space. In contrast, the dark line segment represents the obstacles in the indoor environment.

\subsection{Dataset Evaluation}
Each grid cell of occupancy grid map contains the distributed probability of obstacles and free-space in the actual indoor environment. Fig.~\ref{dataset} (a)–(d) show the created occupancy grid maps in different indoor scenarios. The criteria for evaluating the abnormal occupancy grid maps are as follows:
\begin{itemize}
\item The region in occupancy grid map is inconsistent with the actual environment, such as the repeated mapping and the overlap region, like Fig.~\ref{dataset} (c).
\item The edges of obstacles on the map are unclear, like Fig.~\ref{dataset} (d).
\item The meaningful information is missing in the scanning region of mobile robot, like Fig.~\ref{dataset} (c) and (d).
\end{itemize}

\section{METHOD}
\begin{figure*}[ht]
    \centering
    \includegraphics [width=6.5in]{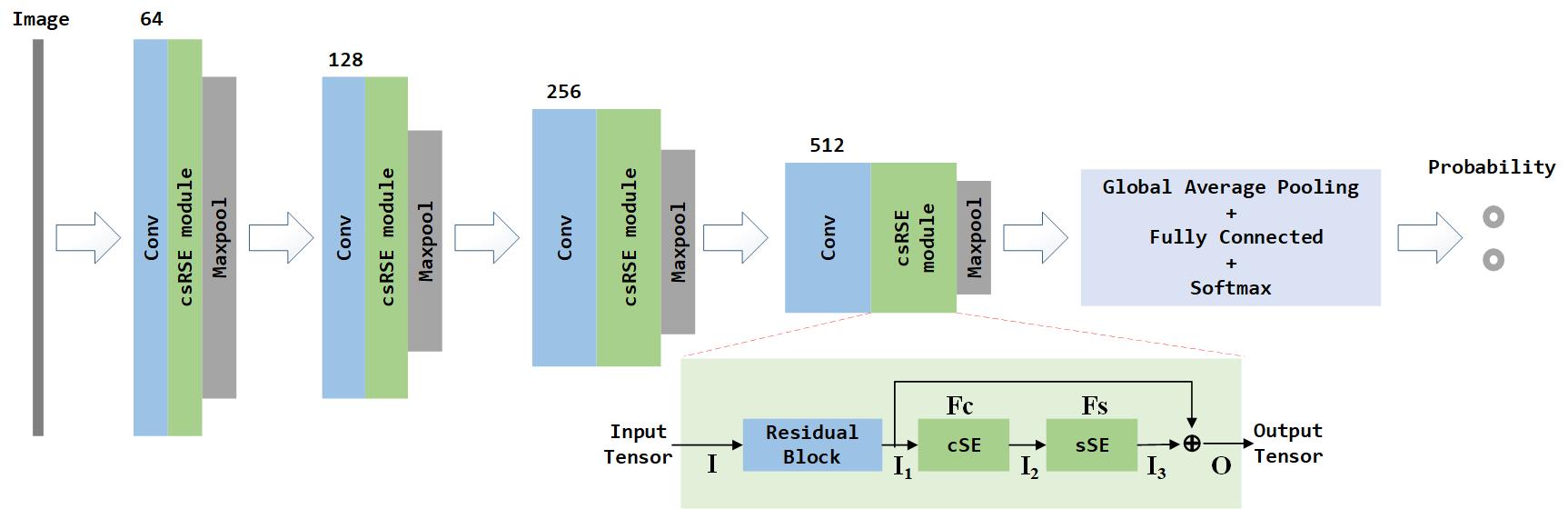}
    \vspace{0mm}
    \caption{The csRSE-integrated network (ResNet32 + csRSE) architecture for occupancy grid  map recognition. The Conv, and csRSE module represent the convolution layer, the channel and spatial Residual Squeeze-and-Excitation (csRSE) module. Given the input feature map $I \in \mathbb{R}^{C \times H \times W}$, through the residual block calculation, we get the feature $I_{1}$. Then through the cSE block and sSE block, it can compute corresponding channel attention map $I_{2}$ and final attention map $I_{3}$. After the element-wise multiplication, we get the refined feature map $O$.}
    \label{architecture}
    \vspace{-3mm}
\end{figure*}

The detailed structure of our attention network is illustrated in Fig.~\ref{architecture}. In the figure, for the input feature map $I \in \mathbb R^{C \times H \times W}$, our attention network computes the channel and spatial attention map $O \in \mathbb R^{C \times H \times W}$. Assuming the input and output dimensions are the same, through channel and spatial Residual SE (csRSE) module, the refined feature map transformation $I_{1} \to O$ can be defined as
\begin{align}
    O=I_{1}+I_{1} \otimes I_{3},
\end{align}
where $\otimes$ denotes element-wise multiplication. Through the residual block calculation and the attention mechanism recalibration, the network can extract global informative features. In our module, after the residual block calculation, we get the feature $I_{1}$, then we compute the channel attention $F_{c}\in \mathbb{R}^{C \times 1 \times 1}$and the spatial attention $F_{s}\in \mathbb{R}^{1 \times H \times W}$ through the cSE block and the sSE block, so the attention map is computed as
\begin{align}
    I_{2} =I_{1} \otimes F_{c}(I_{1}),
\end{align}
\begin{align}
     I_{3} = I_{2} \otimes F_{s}\left(I_{2}\right),
\end{align}
where the $I_{2}$ is output feature map of channel attention. The $I_{3}$ is the final refined output feature map.
The following subsection will describe the details of cSE Block and sSE block.
\subsection{Channel SE Block}
In the cSE block, we recalibrate the inter-channel relationship for the feature response, which involves two steps, spatial squeezation and channel excitation. In the first step, for the feature map $I_{1} \in \mathbb{R}^{C \times H \times W}$, we use the Global Average Pooling (GAP) to squeeze the global information. Then a unique channel vector $V_{c} \in \mathbb{R}^{C \times 1 \times 1}$of each channel is produced by the mean of GAP. 

In the second step, to estimate attention from the $c$-th element of statistic channel vector, we use the Multi-layer Perceptron (MLP) which contains two Fully Connected (FC) layers, the Rectified Linear Units (ReLU) function, and the sigmoid function to capture channel-wise dependencies. The purpose of this MLP is to emphasize the channels with the meaningful information. In short, the output of channel attention is computed as:
\begin{align}
    \nonumber F_{c}(I_{1}) &=M L P(G A P(I_{1})) \\
    &=\sigma\left(W_{2} \delta\left(W_{1} G A P(I_{1})\right)\right),
\end{align}
where $W_{1} \in \mathbb{R}^{\bar{C} \times C}, W_{2} \in \mathbb{R}^{C \times \bar{C}}$ are the weights of the $\mathrm{FC}$ layers, $\bar{C}=\frac{C}{r}, r$ is the reduction ratio ($r$ is set to 16). $\delta$ refers to the Rectified Linear Units (ReLU) function, $\sigma$ refers to the sigmoid function.


\subsection{Spatial SE Block}
In the sSE block, we use the convolutional layers to implement channel squeeze and spatial excitation. Here, it is assumed an alternative representation of the input tensor as $I_{2}$. Four standard convolution layers $W$ are concatenated to produce the spatial attention map $F_{s}(I_{2})=W *I_{2}$. The sigmoid function $\sigma\left(F_{s}(I_{2})\right)$ determines the importance of the specific location across the feature map. Like the previous block, this recalibration process indicates which locations are more meaningful during the training procedure. As a result, the output of spatial attention block can be expressed as
\begin{align}
    \nonumber I_{3} &= \sigma\left(F_{s}(I_{2})\right) \\
    &=\sigma\left(W * I_{2}\right) \\
    \nonumber &=\sigma\left(f_{4}^{1 \times 1}\left(f_{3}^{3 \times 3}\left(f_{2}^{3 \times 3}\left(f_{1}^{1 \times 1}\left(I_{2}\right)\right)\right)\right)\right),
\end{align}
where $\sigma$ refers to the sigmoid function, $*$ denotes the convolution operation. The $f^{1 \times 1}$ and $f^{3 \times 3}$ represents a convolution operation with the filter size of $1 \times 1$ and $3 \times 3$, respectively.

\subsection{Network for Grid Map Recognition}
For the occupancy grid map recognition task, the goal is to improve representation ability of network architecture by using attention mechanism, which can focus on meaningful features and suppress unnecessary ones. As illustrated in Fig.~\ref{architecture} csRSE module, the residual block is designed for aggregating hierarchical features at multiple levels and accelerating convergence. To contain sufficient spatial information of the intermediate layers, the sSE block are stacked after the cSE block for spatial feature extracting. The sSE block extracts the hierarchical features and improves the representation of interests via four convolution operation. All in all, cSE block and sSE block can be regarded as the extracting core, which can aggregate specific global context features of grid maps. These global features are weighted averaged from all areas via a specific attention map to each specific map areas. As attention maps are computed for each areas, the detail heat map of the OGMD dataset are displayed for the specific areas (as shown in Fig.~\ref{visualization}).

Secondly, to extract multi-scale features, the convolution layers and csRSE modules are employed as the core of feature extraction in the proposed network architecture. Through the convolution operations, it can extract multi-scale features by different kernel size of convolution layer. After the convolution layer, we adopt a csRSE module to emphasize meaningful grid map features along two principal dimensions (channel and spatial axes). To achieve the abnormality in occupancy grid maps, we sequentially apply cSE block and sSE block, so that the csRSE modules can learn ‘what’ and ‘where’ to attend in the channel and spatial axes, respectively. As a result, the convolution layers and csRSE modules efficiently helps the information flow within the network by learning which map feature information to emphasize or suppress.

As illustrated in Fig.~\ref{architecture}, there are four convolutional layers, four csRSE modules, one GAP, and one FC layer in the proposed csRSE-integrated network (ResNet32 + csRSE) architecture. At the beginning of proposed network, the input 3-channel occupancy grid map image is transformed into a 64-channel feature map. The ResNet in the proposed csRSE-integrated network is similar to \cite{he2016deep} but use dilated convolutions \cite{yu2015multi}. The stacked block is defined as one consecutive convolution layer, one residual block, one cSE block, and one sSE block. The outputs of two attention blocks are fused to generate high-level contextual features, which will be used to guide the low-level contextual features extracted by the residual block. The residual block and two attention blocks are stacked for blending cross-channel information and spatial hierarchical features together. After each stacked block, the max-pooling is performed and each the filter size gets doubled, which is designed to double the spatial size of feature maps and halve channels. The bottom block is stacked with GAP, FC and softmax layer to compute probabilities of predicted category.

\section{EXPERIMENTS AND RESULTS}
To evaluate the proposed method, we carry out experiments on our OGMD dataset for occupancy grid map recognition. Experimental results demonstrate that the proposed attention network generally outperforms the networks with the SE block and CBAM module, respectively.

\begin{table}[htbp]
  \centering
  \caption{comparisons with methods that have three different attention modules on our OGMD test dataset.}
  \setlength{\tabcolsep}{3mm}
    \begin{tabular}{cccc}
    \toprule
    \textit{\textbf{Methods}} & \multicolumn{1}{p{4.055em}}{\textit{\textbf{Param. (M)}}} & \multicolumn{1}{p{4.055em}}{\textit{\textbf{GFLOPs}}} & \multicolumn{1}{p{4.055em}}{\textit{\textbf{Accuracy (\%)}}} \\
    \midrule
    ResNet20 & 0.25  & 2.019 & 91.27 \\
    \midrule
    ResNet20 + SE & 0.27  & 2.019 & 93.62 \\
    \midrule
    ResNet20 + CBAM & 0.27  & 2.021 & 93.86 \\
    \midrule
    ResNet20 + csRSE(ours) & 0.27  & 2.03  & 94.83 \\
    \midrule
    ResNet32 & 0.45  & 3.418 & 91.43 \\
    \midrule
    ResNet32 + SE & 0.47  & 3.418 & 94.15 \\
    \midrule
    ResNet32 + CBAM & 0.47  & 3.422 & 94.21 \\
    \midrule
    \textbf{ResNet32 + csRSE(ours)} & 0.47  & 3.431 & \textbf{96.23} \\
    \bottomrule
    \end{tabular}%
  \label{accuracy}%
\end{table}%
\subsection{Experimental Setups}
We implement occupancy grid map recognition experiments on our OGMD dataset using the PyTorch framework \cite{paszke2019pytorch}. For training, the image resolution is converted to 224 × 224 and normalized with zero-mean normalization \cite{wang2012comparison}. The entire network is trained end-to-end for 400 epochs by Stochastic Gradient Descent (SGD) with the momentum of 0.9 and a weight decay of 1e-4. The mini-batch size is set to 16. It takes about 8 hours for network to converge on an NVIDIA GTX 2080Ti GPU. We initialize the weights according to the method in \cite{he2015delving} and use binary cross-entropy (BCE) loss function. The formulation of BCE loss is as follows:
\begin{align}
    Loss=- (y_{i} \cdot \log(\hat{y_{i}})+(1-y_{i}) \cdot \log(1-\hat{{y}_{i}})),
\end{align}
where ${y}_{i}$ and $\hat{y_{i}}$ denote the actual label category and the probability of prediction, respectively.

\subsection{Experimental Results}
We conduct experiments to validate the effectiveness of csRSE module by comparing with two popular attention modules (SE, CBAM). In experiments, we evaluate these methods on strong backbones, by replacing two different baseline network (ResNet20, ResNet32) and adding three attention modules (SE, CBAM, and csRSE). From Tab.~\ref{accuracy}, all baseline network methods have achieved measurable competitive results through few calculation parameters. The networks with attention modules get the better accuracy than those without any attention methods. 

Compared with the csRSE-integrated network (ResNet32 + csRSE) and CBAM-integrated network (ResNet32 + CBAM), we can find that the networks with two attention modules get the better performance than the methods with only one channel-wise attention block (ResNet32 + SE). Through the result of csRSE-integrated networks (ResNet32 + csRSE), it is seen that the attention network shows a clear advantage against the other as it gets the better accuracy (96.23\%) on the test dataset. By comparison results on SE block and CBAM, the proposed csRSE attention modules capture the global context information as well as preserve more structural information.

\begin{figure}[ht]
    \centering
    \includegraphics [width=3.4in]{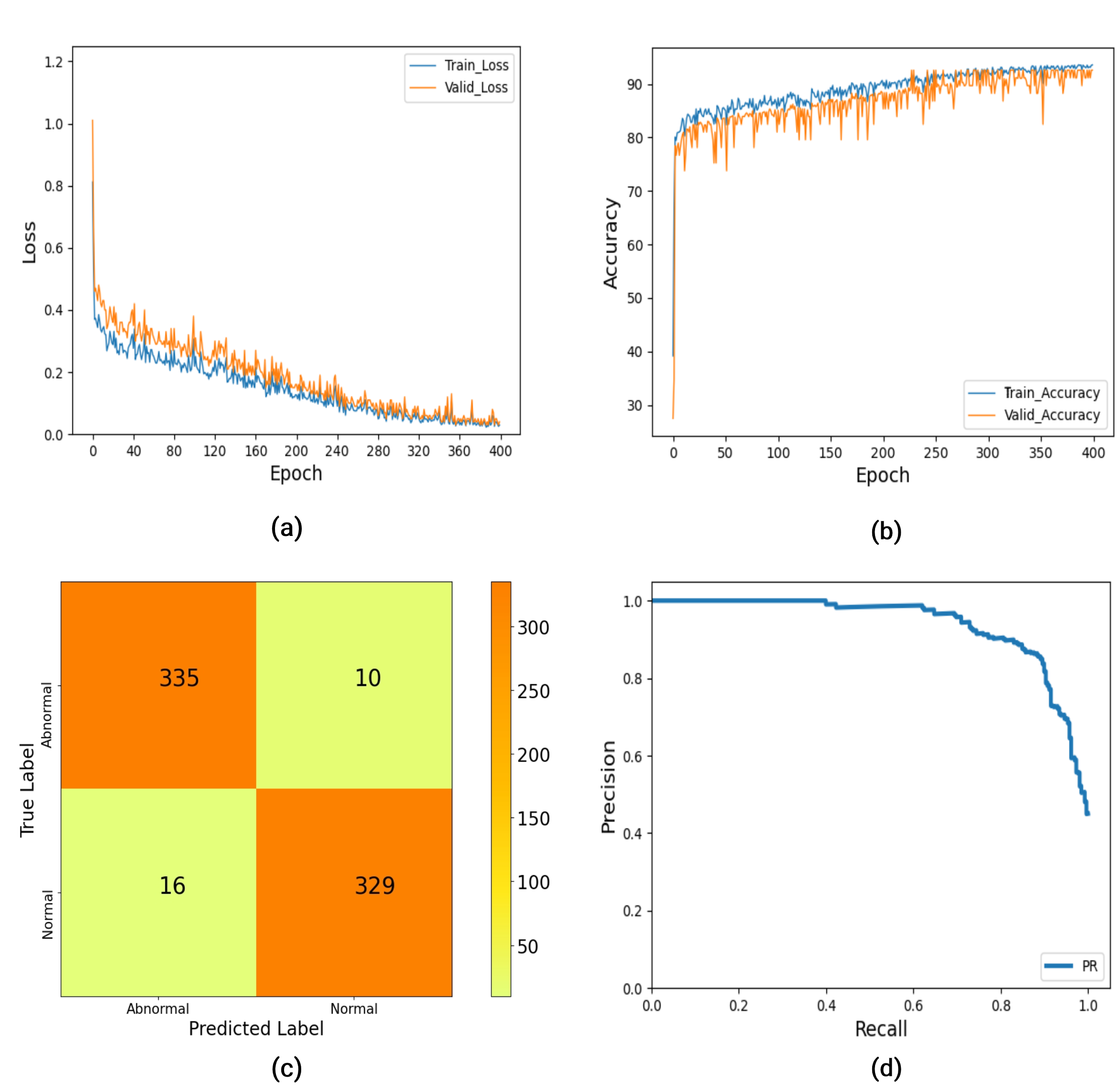}
    \vspace{-5mm}
    \caption{(a) Loss curve, (b) Accuracy curve, (c) Confusion matrix, (d) Precision-Recall curve of experiment results on the test dataset.}
    \label{curve}
    \vspace{-5mm}
\end{figure}

\subsection{Experimental Analysis}
Fig.~\ref{curve} (a) shows the training and validation loss curve of the csRSE-integrated network. We can see that both the training loss and the validation loss are high at the beginning epochs. As the training epoch increases, both the losses remain quite stable. Fig.~\ref{curve} (b) shows the curve of the training accuracy and validation accuracy based on the OGMD dataset. At the beginning, both the training accuracy and validation accuracy are getting stable as the learning rate gradually shrinks. Finally, we can get the best accuracy as the losses decrease.

Fig.~\ref{curve} (c) shows the confusion matrix for our csRSE-integrated network on OGMD test dataset. The y-axis is the true labels and the x-axis is the predicted labels. Overall, the model has achieved a good prediction accuracy of 96.23\%. Fig.~\ref{curve} (d) shows the precision-recall curve of our OGMD test dataset. The y-axis shows the precision value and the x-axis shows the recall value. The experiment results show the actual prediction situation in the occupancy grid map recognition.

\begin{figure*}[t!]
    \centering
    \includegraphics [width=6.3in]{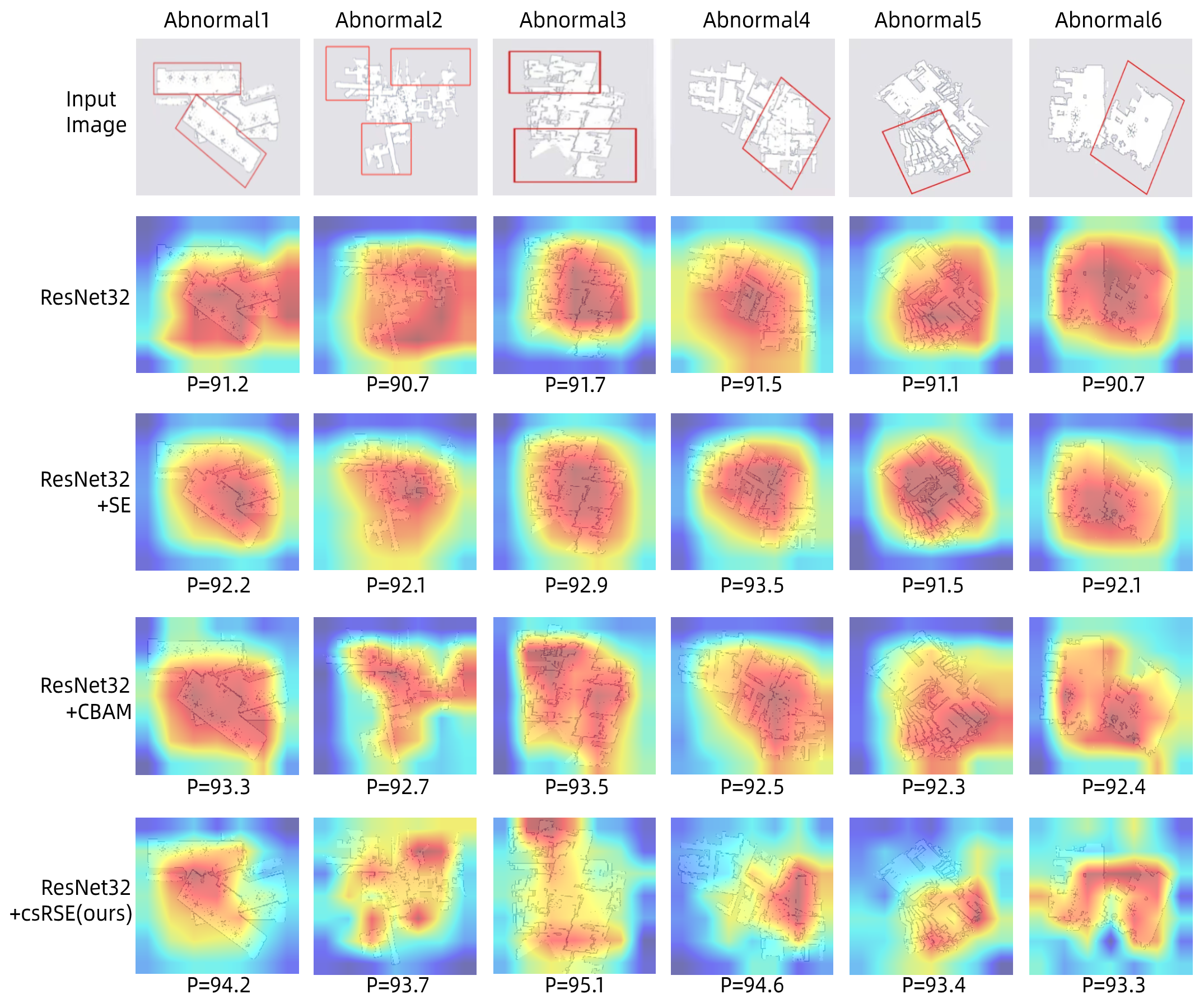}
    \centering
    \vspace{0mm}
    \caption{Visualization results of attention maps (heat maps) on the occupancy grid map recognition. We compare the visualization results of baseline (ResNet32), SE-integrated network (ResNet32 + SE), CBAM-integrated network (ResNet32 + CBAM) and csRSE-integrated network (ResNet32 + csRSE). The regions that marked in the red frame represent the abnormal regions. P denotes the softmax score of each network for the ground-truth class. The higher P means the better classification.}
    \label{visualization}
    \vspace{-5mm}
\end{figure*}
\subsection{Experimental Visualization}
To intuitively verify the effectiveness of the csRSE module, we visualize the attention maps (heat maps) by the Grad-CAM \cite{selvaraju2017grad} on OGMD test dataset. In recent year, Grad-CAM is a visualization method which shows the attended regions of spatial locations in convolutional layers. The Grad-CAM results show attended regions of image clearly. By observing the attended regions, the networks can extract the important feature of image for predicting a class.

In Fig.~\ref{visualization}, we randomly test six abnormal images with different attention networks, including the baseline network (ResNet32), SE-integrated network (ResNet32 + SE), CBAM-integrated network (ResNet32 + CBAM) and our csRSE-integrated network. It can clearly see that the Grad-CAM masks cover attended regions that marked in the red frame represent the abnormal regions. It can be seen that our method is capable of accurately detecting both duplicate maps (e.g., the abnormal1, abnormal2, and abnormal6 input images) and overlapping maps (e.g., the abnormal3, abnormal4, and abnormal5 input images). This is mainly because the global contextual features extracted by the csRSE module can help the network better locate abnormal regions and detect abnormal occupancy grid map. The P denotes final softmax score of networks for the ground-truth class. The higher P has the better classification result, meaning that how this network is making good use of features.

As can be observed, our method can successfully eliminate such occupancy grid maps and detect the abnormal regions. This is mainly contributed by the proposed csRSE attention network, which exploits contextual information in target abnormal regions and provides more texture details for abnormal region localization. Based on the visualization, we conjecture that the proposed csRSE attention network can more effectively detect the abnormal areas from occupancy grid maps and store more global texture details for the occupancy grid map recognition tasks.

\section{CONCLUSION}
In this paper we present a residual neural network using attention mechanism for the abnormal occupancy grid map recognition task. We contribute an OGMD dataset that covers various occupancy grid maps. Then we propose a csRSE attention module, which contains a residual block for producing hierarchical features, followed by both cSE block and sSE block for the sufficient global information from channel-wise and spatial axes. Our attention network is constructed via applying multiple convolution layers and csRSE modules to multi-scale features extraction, which achieves a SOTA prediction accuracy of 96.23\% for the occupancy grid map recognition. In future works, we are going to detect and predict the moving obstacles in the outdoor environment. Moreover, we will apply this attention network to study the problem of global localization for vision based multi-robot SLAM.


\bibliographystyle{IEEEtran}
\bibliography{reference.bib}

\end{document}